\documentclass[pdflatex,sn-mathphys-num]{sn-jnl}

\usepackage{graphicx}%
\usepackage{multirow}%
\usepackage{amsmath,amssymb,amsfonts}%
\usepackage{amsthm}%
\usepackage{mathrsfs}%
\usepackage[title]{appendix}%
\usepackage{xcolor}%
\usepackage{textcomp}%
\usepackage{manyfoot}%
\usepackage{booktabs}%
\usepackage{algorithm}%
\usepackage{algorithmicx}%
\usepackage{algpseudocode}%
\usepackage{listings}%

\theoremstyle{thmstyleone}%
\theoremstyle{thmstyletwo}%

\theoremstyle{thmstylethree}%

\begin{document}

\title[Article Title]{SMiCRM: A Benchmark Dataset of Mechanistic Molecular Images}





\author[1]{\fnm{Ching Ting} \sur{LEUNG}}\email{ctleungaf@connect.ust.hk}
\equalcont{These authors contributed equally to this work.}
\author[1]{\fnm{Yufan} \sur{CHEN}}\email{ychenkv@connect.ust.hk}
\equalcont{These authors contributed equally to this work.}

\author*[1]{\fnm{Hanyu} \sur{GAO}}\email{hanyugao@ust.hk}

\affil[1]{\orgdiv{Department of Chemical and Biological Engineering}, \orgname{Hong Kong University of Science and Technology}, \country{Hong Kong SAR}}





\abstract{Optical chemical structure recognition (OCSR) systems aim to extract the molecular structure information, usually in the form of molecular graph or SMILES, from images of chemical molecules. While many tools have been developed for this purpose, challenges still exist due to different types of noises that might exist in the images. Specifically, we focus on the “arrow-pushing” diagrams, a typical type of chemical images to demonstrate electron flow in mechanistic steps. We present Structural molecular identifier of Molecular images in Chemical Reaction Mechanisms (SMiCRM), a dataset designed to benchmark machine recognition capabilities of chemical molecules with arrow-pushing annotations. Comprising 453 images, it spans a broad array of organic chemical reactions, each illustrated with molecular structures and mechanistic arrows. SMiCRM offers a rich collection of annotated molecule images for enhancing the benchmarking process for OCSR methods. This dataset includes a machine-readable molecular identity for each image as well as mechanistic arrows showing electron flow during chemical reactions. It presents a more authentic and challenging task for testing molecular recognition technologies, and achieving this task can greatly enrich the mechanisitic information in computer-extracted chemical reaction data.}

\keywords{reaction mechanisms, computer vision, dataset, molecular identification}



\maketitle
\maketitle

\section{Objective}

Chemical reaction information is mostly dispersed in different literature in the format of human-readable text and images. Conversion of such loosely organized information into computer-readable databases is essential for further development in reaction recognition and prediction with chem-informatics. Images of different molecules are converted into computer-readable representations of the molecule’s identity using optical chemical structure recognition (OCSR) systems \cite{Rajan2020OSCR}. By 2023, there are several OCSR models targeted for converting molecular images into their computer-readable molecular identities \cite{2009OSRA, Clevert2021Img2Mol, Qian2023MolScribe, Qian2023RxnScribe,Rajan2021Decimer1,RajanDecimer2020Decimer,Xu2022SwinOCSR}. For example, two recent models, DECIMER \cite{RajanDecimer2020Decimer} and MolScribe \cite{Qian2023MolScribe} have been developed to perform OCSR tasks on a variety of datasets, such as 5719 examples from patent grants released by the United States Patent and Trademark Office (USPTO dataset) \cite{marco2015uspto}. ACS dataset is a dataset of molecular images collected from journal articles by previous work \cite{Qian2023MolScribe}. ChemDraw dataset is a synthetic dataset, rendered from 5719 molecules in the USPTO dataset. While there is still room for improvement on the ACS dataset which contains more complexity from real images, in general it can be concluded that both models demonstrated good performances on these datasets.

\begin{figure*}[t]
\centering
\includegraphics[width=0.6\textwidth]{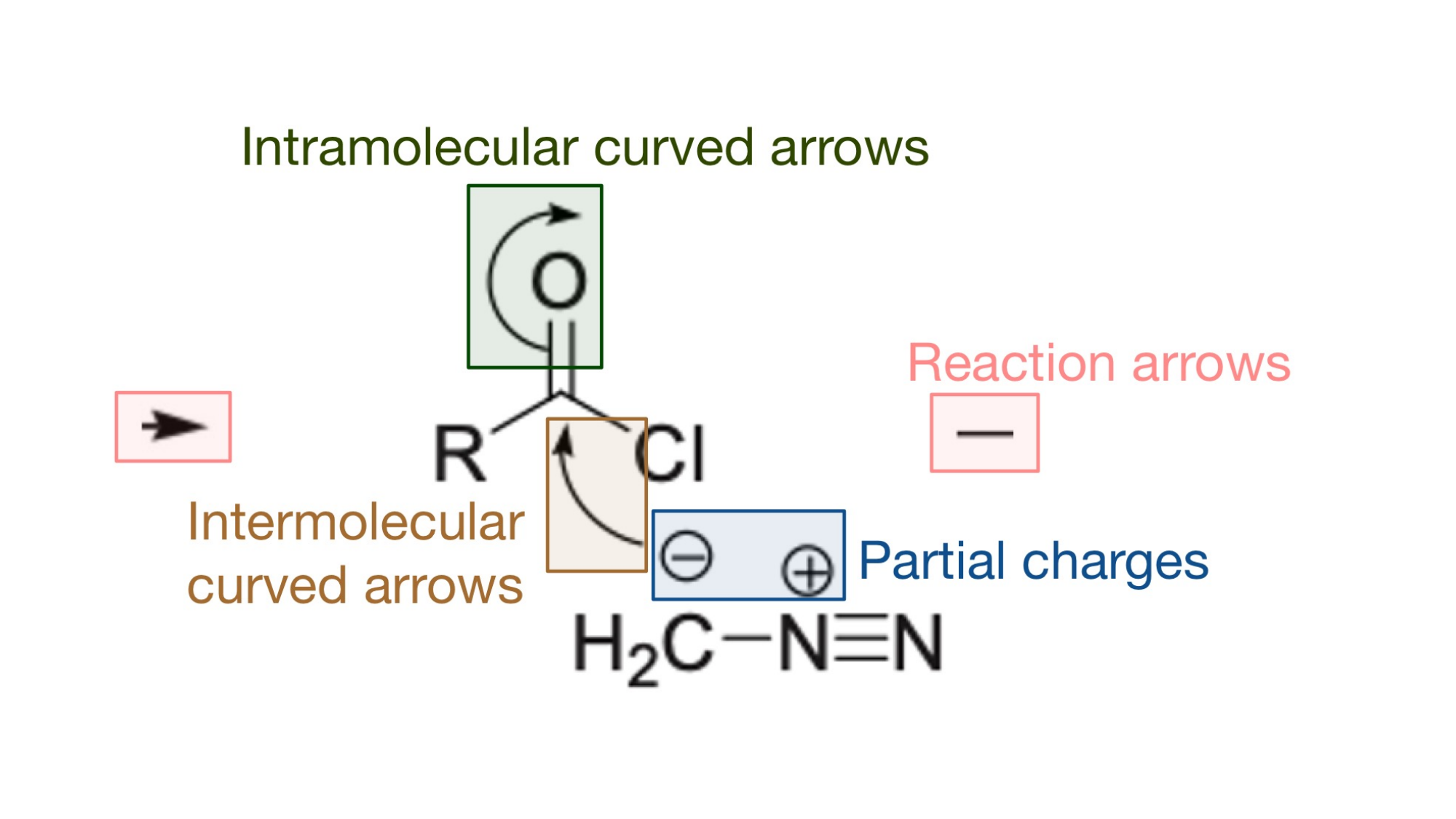}
\caption{Demonstration of a typical noised molecule image in chemical reaction mechanisms. It contains 4 types of contamination: intra- and inter-molecular curved arrows, partial charges and reaction arrows.}
\label{fig:1}
\end{figure*}

While many chemical reaction images only included reacting and product molecules, it is not uncommon for organic chemical reaction images to include additional information such as partial charges and curved arrows indicating electron movements demonstrated in Fig. \ref{fig:1}. This is in fact important information for chemists who read these images and analyze the chemical information in them, and the lack of mechanistic insight is one major criticism that is frequently raised for ML-based reaction prediction models. From a computer vision point of view, this information creates noises that will heavily impact the identification of molecules by OCSR. We tested the performance of some state-of-the-art models on these molecules. As shown in Table \ref{tab:performance}, the current model generally shows a low accuracy in translating mechanistic molecular images into their computer-readable identity, compared to their satisfactory performance on other OCSR tasks. These images are even properly computer drawn, compared to some other datasets that have hand-drawn images, yet the accuracy was much lower that would most probably be attributed to the additional electron-pushing information present in the diagrams. Due to the importance of mechanistic information, we would like to put forward this database to the academic community and call for better solutions.

\begin{table}[t]
  \centering
  \renewcommand\arraystretch{1.5}
  \caption{Result comparisons (\%) of the overall SMILES sequence exact matching accuracy of MiCRM and other existing datasets that have no mechanistic characteristic.}
  \label{tab:performance}
  \begin{tabular}{c  c  c  c  c  c  c} 
 \hline
 Metrics & Method & SMiCRM & USPTO \cite{marco2015uspto} & ACS \cite{Qian2023MolScribe} & ChemDraw \\ [0.5ex] 
\hline\hline

Exact matching of SMILES & DECIMER \cite{Rajan2021Decimer1} & 7.5 & 74.2 & 46.5 & 86.1\\
 & MolScribe \cite{Qian2023MolScribe} & 10.6 & 91.9 & 69.8 & 93.8 \\
  \midrule
 Tanimoto Smiliarity & DECIMER & 19.7 & 91.3 & 61.3 & 99.4 \\
 & MolScribe & 45.1 & 97.2 & 76.2 & 97.5\\[0.5ex]
 
 \hline
 
\end{tabular}
  
\end{table}


\section{Data Description}
The dataset consists of 453 PNG images of molecules from chemical reaction mechanisms. 17 images are obtained from reaction image datasets created by previous work \cite{Qian2023RxnScribe}, and cropped only to retain the molecular image. The remaining 436 images are captured and extracted from reaction mechanisms from a collection of named chemical reactions \cite{Li2009reactions}. The images are then labeled with their corresponding Simplified Molecular Input Line Entry System (SMILES) \cite{Weininger1988SMILES} and recorded in our dataset as well as individual structural data files (SDF).

\subsection{Curation}
\begin{figure*}[t]
\centering
\includegraphics[width=0.8\textwidth]{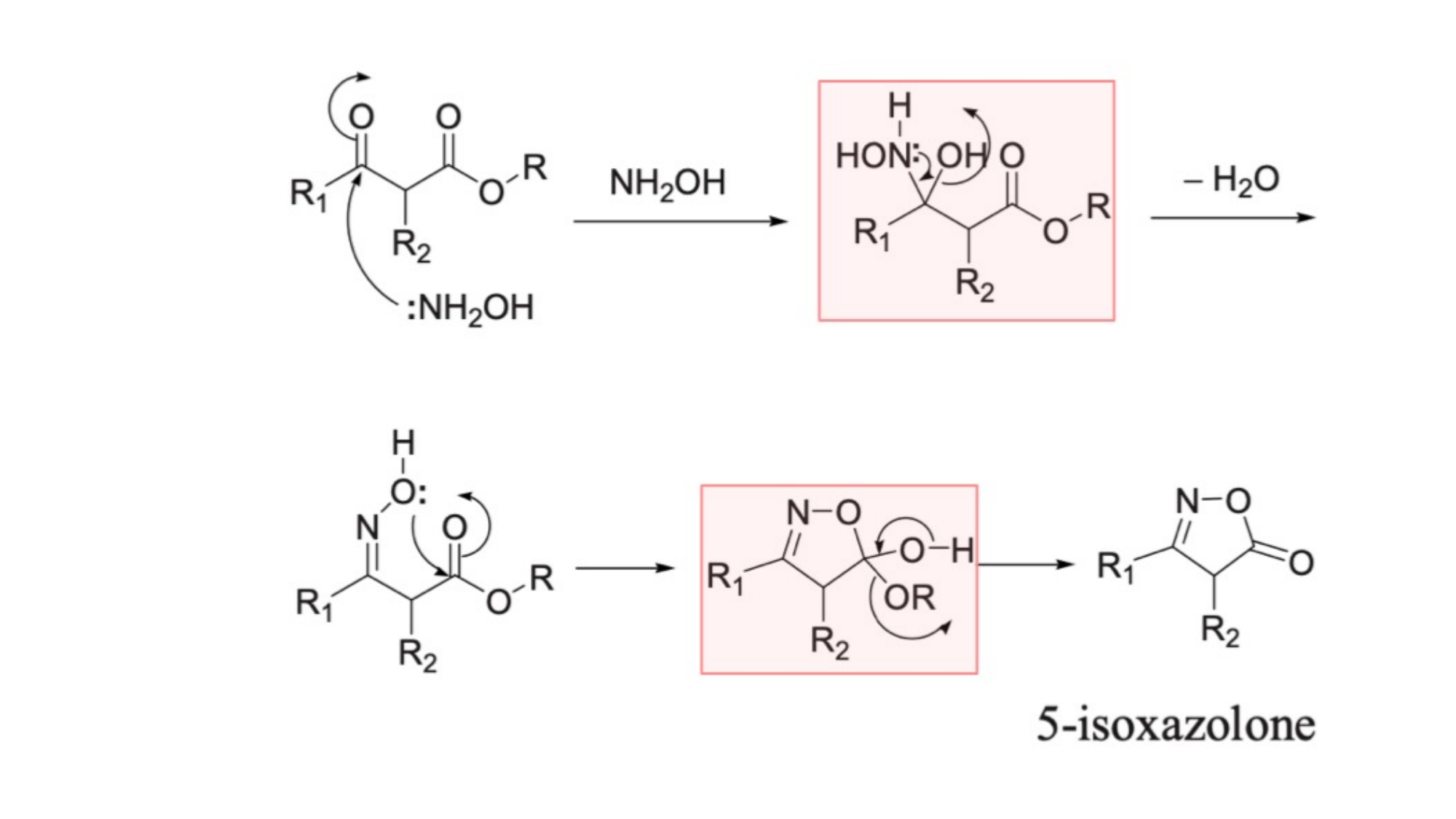}
\caption{Only the highlighted molecule is documented for this reaction: The second molecule is an aliphatic molecule containing more functional groups, the forth one is an aromatic molecule containing more functional groups. Both molecules are chosen as they have curved arrows. While the first and the forth molecule are relatively simple in structure, and the last molecule does not have any curved arrows on.}
\label{fig:2}
\end{figure*}

\begin{figure*}[t]
\centering
\includegraphics[width=0.9\textwidth]{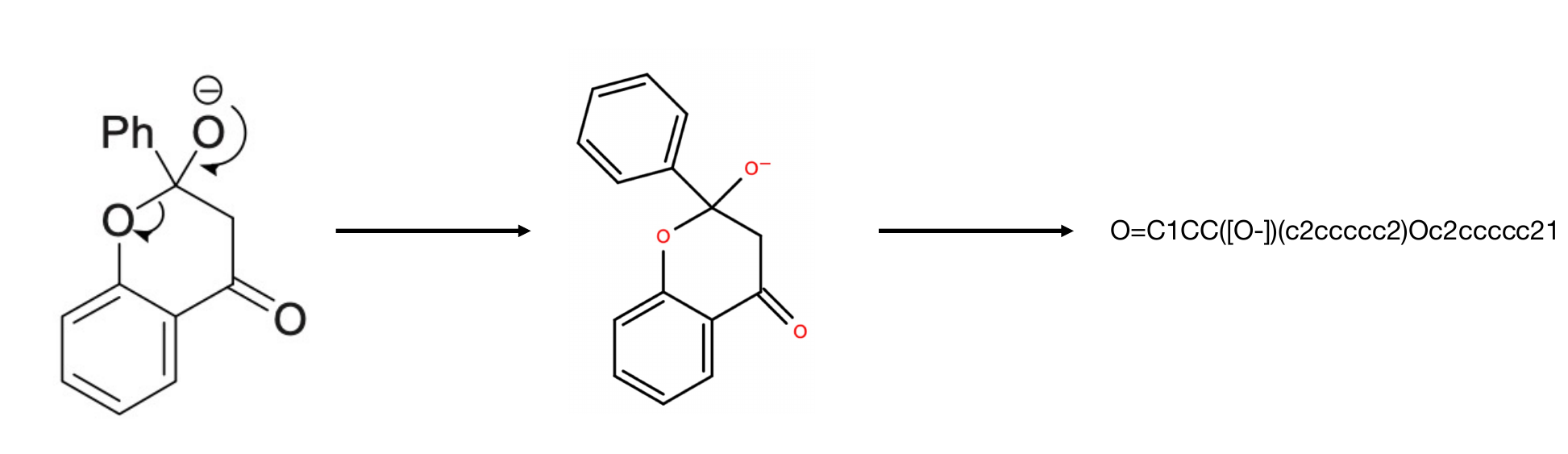}
\caption{Demonstration of extraction of molecular information from images. Identities of abbreviations are revealed, curved arrows are ignored, and partial charges are included in the molecule's identity.}
\label{fig:3}
\end{figure*}

The dataset consists of a variety of images from different reaction mechanisms. Going through all the reactions, only molecules with mechanistic features such as curved arrows and partial charges are selected. In addition, only one to two mechanistic molecular images are selected per reaction mechanism to preserve the uniqueness of every image, examples of molecules being chosen demonstrated in Fig. \ref{fig:2}. Each image is then manually drawn as a molecular graph by an open-source computer-aid synthesis planning (CASP) tool, ASKCOS \cite{ASKCOS} and converted to canonical SMILES shown in Fig. \ref{fig:3}. The canonical SMILES obtained are then converted using RDKit \cite{landrum2013rdkit} into separate SDFs.

\subsection{FAIR-ification}
To ensure the dataset meets FAIR principles of being findable, accessible, interoperable, and reusable \cite{Jacobsen2020FAIR}, it was deposited in the open repository Zenodo and assigned a DOI for identification. Zenodo enables global access and integration with development platforms like GitHub. Structures are represented as portable PNG images and canonical SMILES formats compatible with cheminformatics tools. The CC-BY 4.0 license allows free reuse, redistribution, and modification of the data, provided appropriate attribution is given. Commercial and non-commercial applications are permitted without additional constraints. By publishing in Zenodo with standardized data formats and an open license, the dataset content can be easily discovered, shared, and integrated and its reuse is well-defined, facilitating the goals of findability, accessibility, interoperability, and reusability. The CC-BY terms further maximize opportunities for the data to enable novel research and advance the field through broad dissemination and adaptation by the scientific community.
\section{Limitations}
This benchmark dataset can be freely used and reused without limitations for research and evaluation purposes. It is designed as a standardized test set for assessing the performance of OCSR tools on diverse chemical structures commonly found in organic reaction schemes.
The dataset contains a wide range of molecular images that could appear in chemical reaction mechanisms. Each image is accompanied by its canonical SMILES representation for standardized evaluation. With around 450 images, it provides a challenging set for measuring capabilities in visual understanding of chemistry.

However, due to its size, we do not recommend utilizing this dataset alone for training machine learning models. Its intended role is to fairly and independently benchmark existing or newly developed OCSR systems, avoiding potential overfitting to this limited collection of examples. Researchers are encouraged to use it exclusively to gauge tool performance rather than parameter fitting. Advancing tools through this standardized evaluation will contribute to progress in machine reading of complex chemical diagrams.

\bmhead{List of abbreviations} OCSR: Optical Chemical Structure Recognition; SMILES: Simplifed Molecular-Input Line-entry System; CC: Creative commons; DOI: Digital object identifier; FAIR: Findable, accessible, interoperable, and reusable; CASP: Computer-Aid Synthesis Planning; PDF: Portable document format; PNG: Portable network graphics; SDF: structural data file. 

\bmhead{Availability of data and materials}
The dataset of this article is available in \url{https://doi.org/10.5281/zenodo.11060696}.

\section*{Declarations}

\bmhead{Competing interests}
The authors declare no competing financial interest.

\bmhead{Funding}
HKUST (Project No. R9251, Z1269)

\bmhead{Authors' contributions}
CTL wrote the main manuscript. CTL and YC curated the database collaboratively. CTL, YC and HG designed the method collaboratively. HG supervised the work. All authors read and approved the final manuscript.

\bibliography{sn-bibliography}

\end{document}